\documentclass[a4paper,conference]{IEEEtran}
\usepackage[utf8]{inputenc}
\usepackage[english]{babel}
\usepackage{paralist}
\usepackage{csquotes}
\usepackage[style=ieee, mincitenames=1, maxcitenames=1]{biblatex}
\usepackage{url}
\usepackage{hyperref}
\usepackage{graphicx}
\usepackage{caption}
\usepackage[frozencache,cachedir=.]{minted}
\usepackage{subcaption}
\usepackage{booktabs}
\newcommand{\bftab}{\fontseries{b}\selectfont}
\usepackage{graphicx}
\usepackage[binary-units=true]{siunitx}
\DeclareSIUnit\px{px}

\usepackage[acronym]{glossaries}
\newacronym{dr}{DR}{Domain Randomization}
\newacronym{map}{mAP}{mean Average Precision}
\newacronym{iou}{IoU}{Intersection over Union}

\newif\ifreview
\reviewfalse

\begin{document}
\title{BlendTorch: A Real-Time, Adaptive \\ Domain Randomization Library}

\ifreview
    \author{\IEEEauthorblockN{--Hidden For Peer Review--}}
\else
    \author{
        \IEEEauthorblockN{
            Christoph Heindl\IEEEauthorrefmark{1},
            Lukas Brunner\IEEEauthorrefmark{1},
            Sebastian Zambal\IEEEauthorrefmark{1} and
            Josef Scharinger\IEEEauthorrefmark{2}}
        \IEEEauthorblockA{\IEEEauthorrefmark{1} Visual Computing, Profactor GmbH, Austria, \texttt{first.last@profactor.at}}
        \IEEEauthorblockA{\IEEEauthorrefmark{2} Computational Perception, JKU, Austria, \texttt{josef.scharinger@jku.at}}
    }
\fi

\maketitle

\begin{abstract}
Solving complex computer vision tasks by deep learning techniques relies on large amounts of (supervised) image data, typically unavailable in industrial environments. The lack of training data starts to impede the successful transfer of state-of-the-art methods in computer vision to industrial applications. We introduce BlendTorch, an adaptive \acrfull{dr} library, to help creating infinite streams of synthetic training data. BlendTorch generates data by massively randomizing low-fidelity simulations and takes care of distributing artificial training data for model learning in real-time. We show that models trained with BlendTorch repeatedly perform better in an industrial object detection task than those trained on real or photo-realistic datasets.
\end{abstract}

%
\IEEEpeerreviewmaketitle

\section{Introduction}
Recent advances in computer vision depend extensively on deep learning techniques. With sufficient modeling capacity and enough (labeled) domain datasets, deeply learned models often outperform conventional vision pipelines~\cite{krizhevsky2012imagenet,simonyan2014very,he2016deep}. However, providing large enough datasets is challenging within industrial applications for several reasons: \begin{inparaenum}[a)]
\item costly and error-prone manual annotations,
\item the odds of observing rare events are low, and
\item a combinatorial data explosion as vision tasks become increasingly complex.
\end{inparaenum}
If high capacity models are trained despite low data quality, the likelihood of overfitting increases, resulting in reduced robustness in industrial applications~\cite{shorten2019survey}.

In this work\footnote{Supported by MEDUSA, Leitprojekt Medizintechnik}, we focus on generating artificial training images and annotations through computer simulations. Training models in simulations promises annotated data without limits, but the discrepancy between the distribution of training and real data often leads to poorly generalizing models~\cite{tremblay2018training}. Increasing photo realism and massive randomization of low-fidelity simulations (\acrlong{dr}) are two popular and contrary strategies to minimize the distributional mismatch. Recent frameworks focus on photorealism, but do not address the specifics of massive randomization such as: online rendering capabilities and feedback channels between training and simulation.

\begin{figure}[t]
    \centering
    \includegraphics[width=\columnwidth]{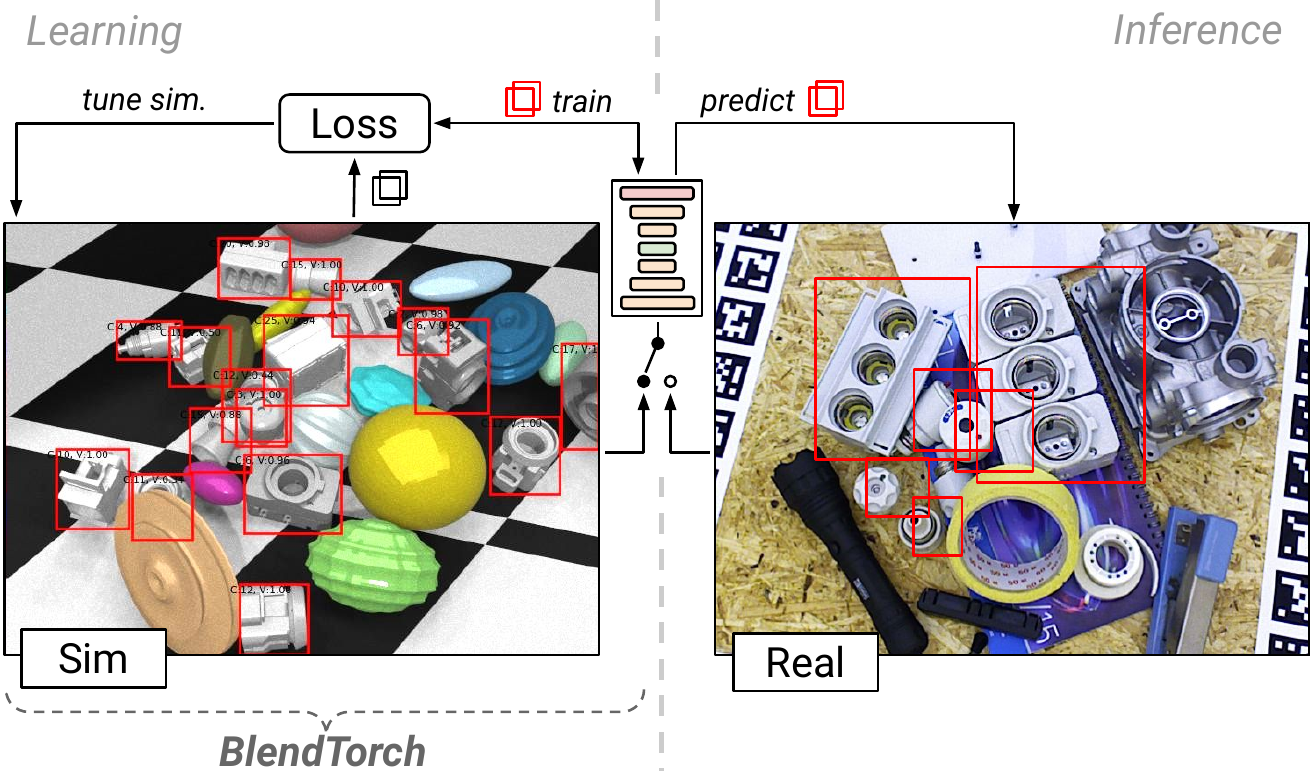}
    \caption{We introduce BlendTorch, a real-time adaptive \acrlong{dr} library, for neural network training in simulated environments. We show, networks trained with BlendTorch outperform identical models trained with photo-realistic or even real datasets on the same object detection task.}
    \label{fig:intro}
\end{figure}

We introduce BlendTorch\footnote{\ifreview \url{http://after-review.com}\else \url{https://github.com/cheind/pytorch-blender}\fi}, a general purpose open-source image synthesis framework for adaptive, real-time \acrfull{dr} written in Python. BlendTorch integrates probabilistic scene composition, physically plausible real-time rendering, distributed data streaming, and bidirectional communication. Our framework integrates the modelling and rendering strengths of Blender~\cite{blender2020} with the high-performance deep learning capabilities of PyTorch~\cite{paszke2019pytorch}. We successfully apply BlendTorch to the task of learning to detect industrial objects without access to real images during training (see Figure~\ref{fig:intro}). We demonstrate that data generation and model training can be done in a single online sweep, reducing the time required to ramp-up neural networks significantly. Our approach not only outperforms photo-realistic datasets on the same perception task, we also show that \acrshort{dr} surpasses the detection performance compared to a real image data set.

\subsection{Related Work}
Using synthetically rendered images for training supervised machine learning tasks has been studied before. \citeauthor{tobin2017domain}~\cite{tobin2017domain} as well as \citeauthor{sadeghi2016cad2rl}~\cite{sadeghi2016cad2rl} introduced uniform \acrlong{dr}, in which image data is synthesized low-fidelity simulations, whose simulation aspects are massively randomized. These earlier \acrshort{dr} approaches were tailored to a specific application, which severely limits their reusability in other environments. BlendTorch is based on the idea of \acrshort{dr}, but generalizes to arbitrary applications.

\begin{figure*}[t!]
    \centering
    \includegraphics[width=0.9\textwidth]{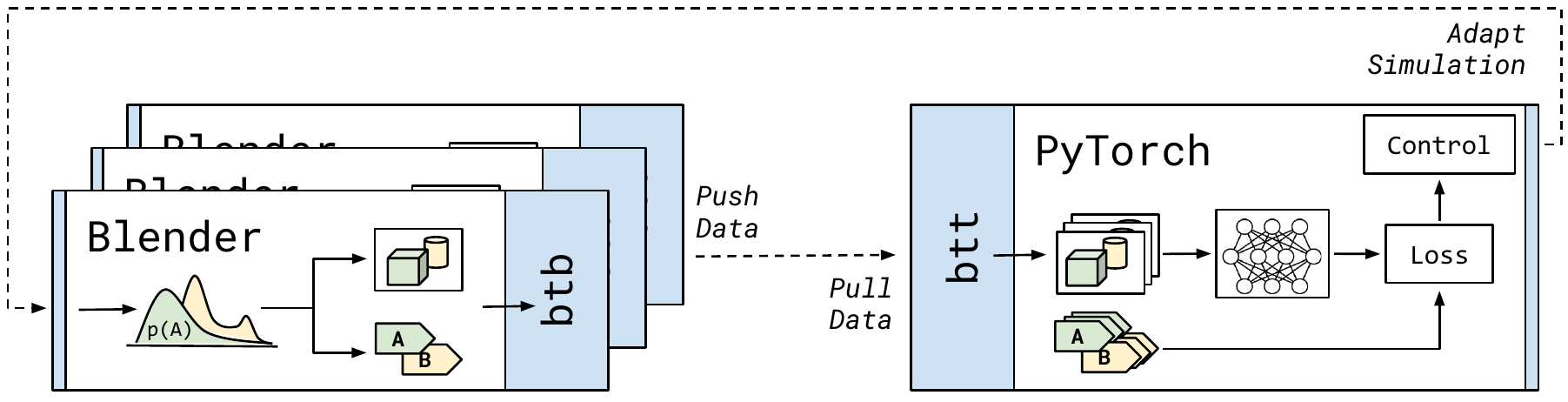}
    \caption{BlendTorch overview. BlendTorch combines probabilistic (supervised) image generation in Blender~\cite{blender2020} (left) with deep learning in PyTorch~\cite{paszke2019pytorch} (right) via a programmatic Python interface. Information is exchanged via a scalable network library (dotted lines). Implementation details are encapsulated in two Python subpackages \texttt{blendtorch.btb} and \texttt{bendtorch.btt}. Additional feedback capabilities allow the simulation adapt to current training needs.}
    \label{fig:arch}
\end{figure*}

Recently, general purpose frameworks focusing on generating photo-realistic images have been introduced~\cite{denningerblenderproc, to2018ndds, schwarz2020stillleben}. Compared to BlendTorch, these frameworks are not real-time capable and also lack a principled way to communicate information from model training back into simulation. The work most closely related to ours is BlenderProc ~\cite{denningerblenderproc}, since we share the idea of using Blender for modelling and rendering purposes. BlenderProc utilizes a non real-time physically-based path tracer to generate photo-realistic images, but lacks real-time support and feedback capabilities offered by BlendTorch. BlenderProc focuses on a configuration file based scene generation, while BlendTorch offers a more flexible programming interface.

Enabling adaptive simulations through training feedback was introduced in \citeauthor{robotpose_etfa2019_cheind}~\cite{robotpose_etfa2019_cheind} for the specialized task of robot keypoint detection. BlendTorch generalizes this idea to arbitrary applications and to real-time rendering.

\subsection{Contributions}
This paper offers the following contributions
\begin{enumerate}
    \item BlendTorch, an adaptive, open-source, real-time domain randomization library that seamlessly connects modelling, rendering and learning aspects.
    \item A comprehensive industrial object detection experiment that highlights the benefits of \acrshort{dr} over photo-realistic and even real training datasets.
\end{enumerate}

\section{Design Principles}
BlendTorch weaves several ideas into a design that enables practitioners and scientists to rapidly realize and test novel \acrshort{dr} concepts.

\begin{LaTeXdescription}
  \item[Reuse and connect.] Training neural networks in simulation using \acrshort{dr} requires several specialized software modules. For a successful experiment, modeling and rendering tools as well as powerful learning libraries for deep learning are required. In the recent past, excellent open-source frameworks for the aforementioned purposes have emerged independently from each other. However, these tools are not interconnected and a basic framework for their online interaction is missing. BlendTorch aims to bring these separate worlds together as seamlessly as possible without losing the benefits of either software component.
  
  \item[Real-time computing.] As the complexity of visual tasks increases, the combinatorial scene variety also increases exponentially. Several applications of \acrshort{dr} separate the simulation from the actual learning process, because the slow image generation stalls model learning. However, offline data generation suffers from the following shortcomings: constantly growing storage requirements and the missing possibility of online simulation adaptation. BlendTorch is designed to provide real-time, distributed integration between simulation and learning that is fast enough not to impede learning progress.
  
  \item[Adaptability.] The ability to adapt the simulation parameters during model training has already proven to be beneficial~\cite{robotpose_etfa2019_cheind} before. Adaptability allows the simulation to synchronize with the evolving requirements of the learning process in order to learn more efficiently. The meaning of adaptability is application-dependent and ranges from adjusting the level of simulation difficulty to the generation of adversarial model examples. BlendTorch is designed with generic bidirectional communication in mind, allowing application-specific workflows to be implemented quickly.
\end{LaTeXdescription}

\section{Architecture}
BlendTorch connects the modeling and rendering strengths of Blender~\cite{blender2020} and the deep learning capabilities of PyTorch~\cite{paszke2019pytorch} as depicted in Figure~\ref{fig:arch}. Our architecture considers data generation to be a bidirectional exchange of information, which contrasts conventional offline, one-way data generation. Our perspective enables BlendTorch to support scenarios that go beyond pure data generation, including adaptive domain randomization and reinforcement learning applications. 

To seamlessly distribute rendering and learning across machine boundaries we utilize ZeroMQ~\cite{hintjens2013zeromq} and split BlendTorch into two distinctive sub-packages that exchange information via ZMQ: \texttt{blendtorch.btb} and \texttt{bendtorch.btt}, providing the Blender and PyTorch views on BlendTorch.

\begin{figure*}[t!]
    \captionsetup[subfigure]{labelformat=empty}
    \centering
    \begin{subfigure}[t][12cm]{0.9\columnwidth}
        \begin{subfigure}[t]{\textwidth}
            \centering
            \caption{Training \texttt{train.py}}
            \begin{minted}[obeytabs=true,tabsize=2,fontsize=\footnotesize,frame=lines]{python}
from torch.utils import data
import blendtorch.btt as btt

def main():
	largs = dict(
		scene='cube.blend',
		script='cube.blend.py',
		num_instances=2,
		named_sockets=['DATA'],
	)
	# Launch Blender locally
	with btt.BlenderLauncher(**largs) as bl:
		# Create remote dataset
		addr = bl.launch_info.addresses['DATA']
		ds = btt.RemoteIterableDataset(addr)
		dl = data.DataLoader(ds, batch_size=4)
		
		for item in dl:
			img, xy = item['image'], item['xy']
			print('Received', img.shape, xy.shape)
			# train with item ...

main()
            \end{minted}
        \end{subfigure}
        \vfill
        \begin{subfigure}[b]{\textwidth}
            \centering
            \includegraphics[width=\textwidth]{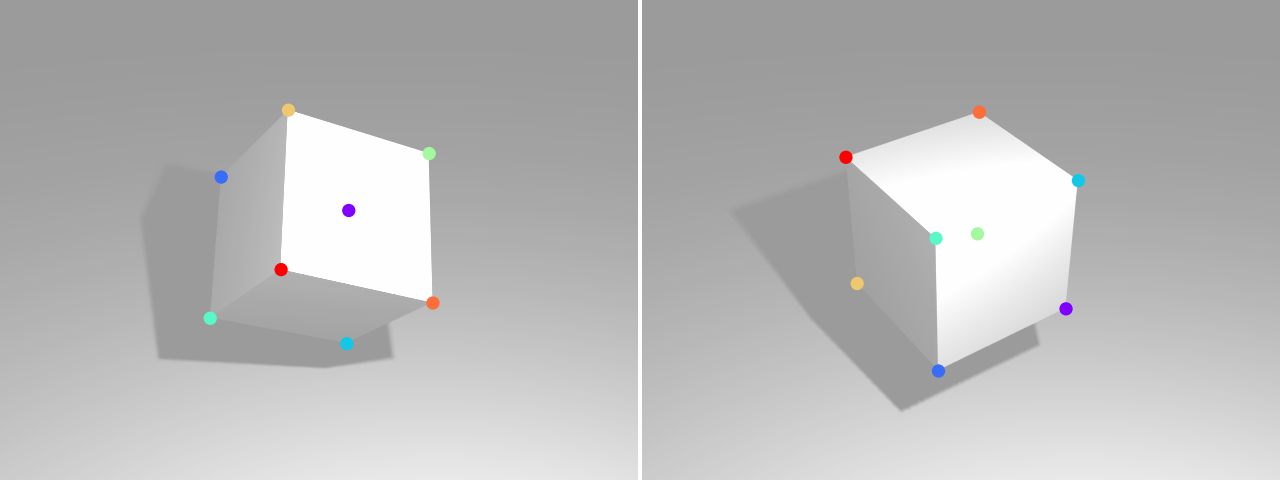}
            \caption{Received images superimposed with annotations.}
        \end{subfigure}
    \end{subfigure}
    \hfill
    \begin{subfigure}[t][12cm]{0.9\columnwidth}
        \centering
        \caption{Simulation \texttt{cube.blend.py}}
        \begin{minted}[obeytabs=true,tabsize=2,fontsize=\footnotesize,frame=lines]{python}
import bpy
from numpy.random import uniform
import blendtorch.btb as btb

def main():	
	# Blender object
	cube = bpy.data.objects["Cube"]
	
	def pre_frame():
		# Randomize
		cube.rotation_euler = uniform(0,3.14,3)
		
	def post_frame():
		# Publish image + annotations
		pub.publish(
			image=off.render(), 
			xy=off.camera.object_to_pixel(cube)
		)

	# Setup data publishing
	btargs, _ = btb.parse_blendtorch_args()
	pub = btb.DataPublisher(
		btargs.btsockets['DATA'], 
		btargs.btid)
		
	# Setup rendering using def. camera
	off = btb.OffScreenRenderer(mode='rgb')
	
	# Setup animation and callbacks
	anim = btb.AnimationController()
	anim.pre_frame.add(pre_frame)
	anim.post_frame.add(post_frame)
	anim.play()

main()
        \end{minted}
    \end{subfigure}
    \caption{Minimal BlendTorch example. The train script \texttt{train.py} (top-left) launches multiple Blender instances to simulate a simple scene driven by a randomization script \texttt{cube.blend.py}. The training then awaits batches of images and annotations within the main training loop. The simulation script \texttt{cube.blend.py} (top-right) randomizes the properties of a cube and publishes color images along with corner annotations. Color images and superimposed annotations are shown in the bottom-left.}
    \label{fig:minimalexample}
\end{figure*}

A typical data generation task for supervised machine learning is setup in BlendTorch as follows. First, the training procedure launches and maintains one or more Blender instances using \texttt{btt.BlenderLauncher}. Each Blender instance will be instructed to run a particular scene and randomization script. Next, the training procedure creates a \texttt{btt.RemoteIterableDataset} to listen for incoming network messages from Blender instances. BlendTorch uses a pipeline pattern that supports multiple data producers and workers employing a fair data queuing policy. It is guaranteed that only one PyTorch worker receives a particular message and no message is lost, but the order in which it is received is not guaranteed. To avoid out-of-memory situations, the simulation processes will be temporarily stalled in case the training is not capable to catch up with data generation. Within every Blender process, the randomization script instantiates a \texttt{btb.DataPublisher} to distribute data messages. This script registers the necessary animation hooks. Typically, randomization occurs in \texttt{pre-frame} callbacks, while images are rendered in \texttt{post-frame} callbacks. Figure~\ref{fig:minimalexample} illustrates these concepts along with a minimal working example.

\section{Experiments}
\begin{figure}[h]
    \centering
    \includegraphics[width=\columnwidth]{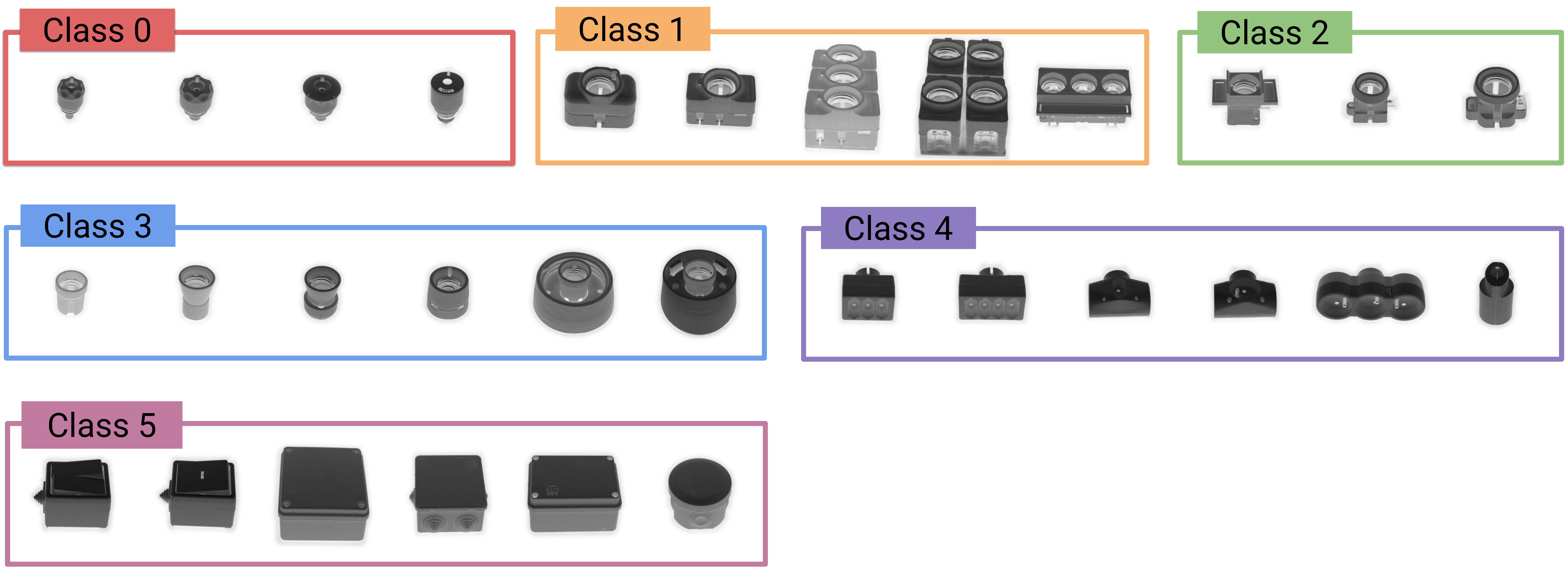}
    \caption{T-Less dataset objects. 30 individual objects re-grouped into 6 super-classes in this work. Displayed in false colors for better visibility.}
    \label{fig:tlessclasses}
\end{figure}

We evaluate BlendTorch by studying its performance within the context of an industrial 2D object detection task. For the following experiments we choose the T-Less dataset~\cite{hodan2017t} which originally consists of 30 industrial objects with no significant texture or discriminative color. For reasons of presentation, we re-group the 30 classes into 6 super-groups as depicted in Figure~\ref{fig:tlessclasses}. T-Less constitutes a solid basis for comparative experiments, since real and synthetic photo-realistic images are already available for training and testing.

Our evaluation methodology can be summarized as follows: We train the same state-of-the-art object detection neural network by varying only the training dataset, but keeping all other hyper-parameters fixed. We then evaluate each resulting model on the same test dataset using the \acrfull{map}~\cite{everingham2010pascal} metric that combines localization and classification performance. To avoid biases due to random model initialization, we repeat model learning multiple times for each dataset.

\subsection{Datasets}
\label{sec:datasets}
\begin{figure}
    \centering
    \begin{subfigure}{0.49\columnwidth}
        \includegraphics[width=\textwidth]{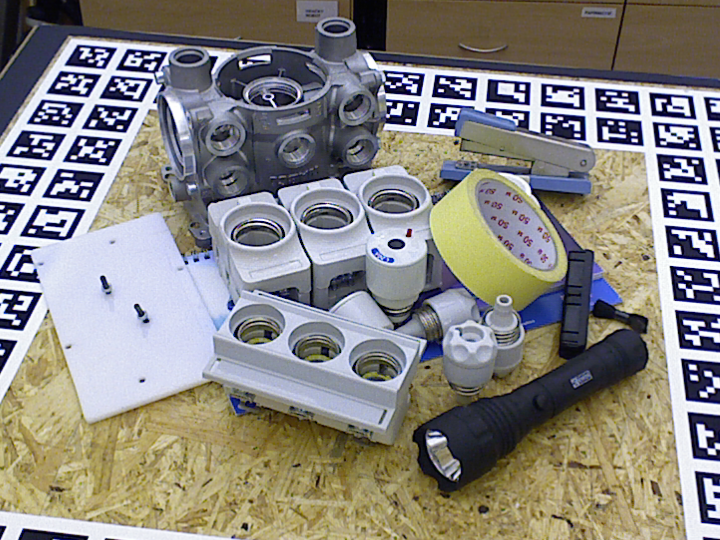}
        \caption{Kinect/BOP dataset sample.}
    \end{subfigure}
    \hfill
    \begin{subfigure}{0.49\columnwidth}
        \includegraphics[width=\textwidth]{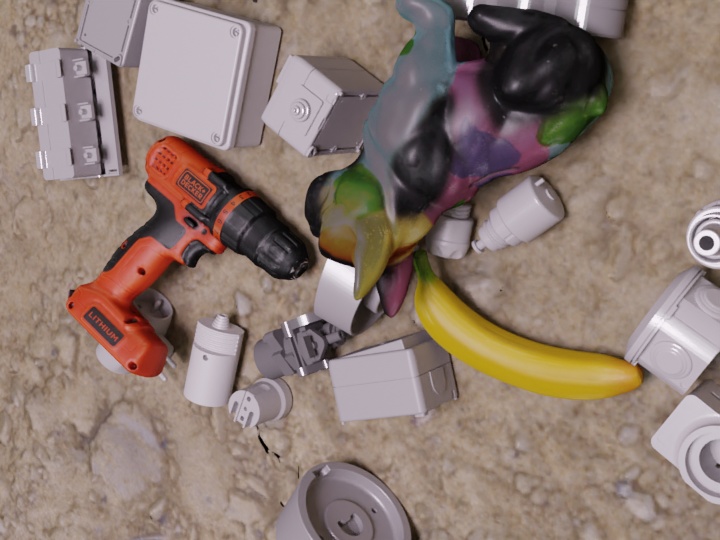}
        \caption{PBR dataset sample.}
    \end{subfigure}
    \begin{subfigure}{\columnwidth}
        \includegraphics[width=\textwidth]{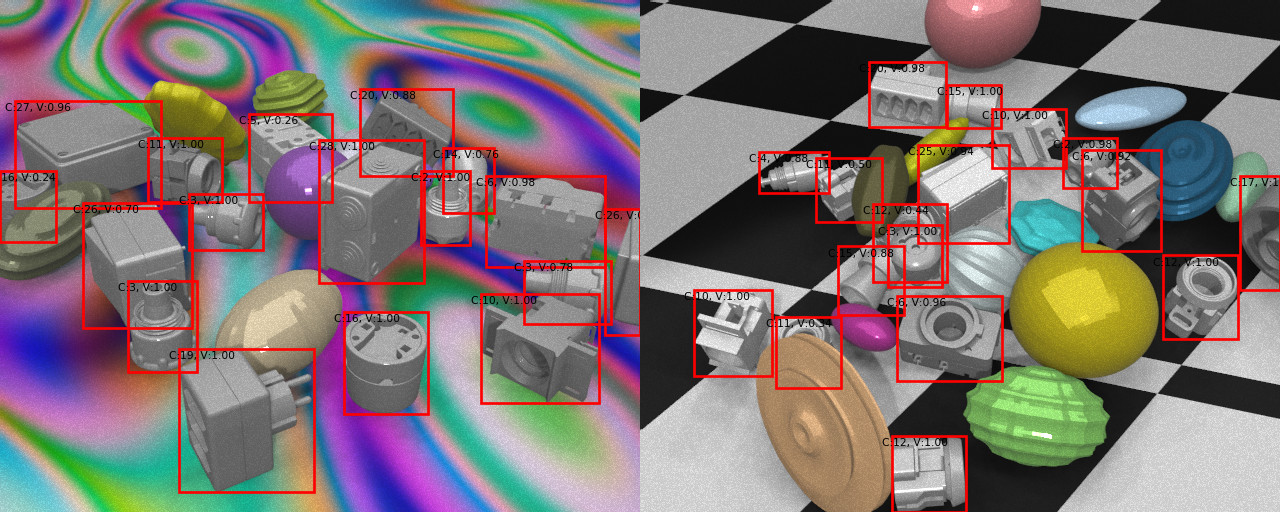}
        \caption{BlendTorch samples with annotated bounding boxes and occluders.}
        \label{fig:datasetsamples:btorch}
    \end{subfigure}
    \caption{Samples from datasets used throughout our work. Notice that PBR gives a much more realistic impression of the scene than BlendTorch using \acrshort{dr}. Yet, we show \acrshort{dr} is simpler to implement, generates images faster, and yields a better performing model.}
    \label{fig:datasetsamples}
\end{figure}

Throughout our evaluation we distinguish four color image T-Less datasets that we describe next.
\begin{LaTeXdescription}
  \item[RealKinect] A real image dataset based on T-Less images taken with a Kinect camera. It consists of $10^{4}$ images grouped into 20 scenes with 500 images per scene. Images are taken in a structured way by sampling positions from a hemisphere. Each scene includes a variable number of occluders.
  \item[PBR] Is a publicly available photo-realistic synthetic image dataset generated by BlenderProc~\cite{denningerblenderproc} in an offline step. It consists of $5 \times 10^{4}$ images taken from random camera positions. PBR uses non parametric occluders which are are inserted randomly.
  \item[BlendTorch] This dataset corresponds to synthetic data generated by \acrlong{dr} BlendTorch. The dataset consists of $5 \times 10^{4}$ images, whose generation details are given in Section~\ref{sec:btorchdataset}.
  \item[BOP] A real image dataset of T-Less corresponding to the test dataset of the BOP Challenge 2020~\cite{hodan2020bop} taken with a PrimeSense camera. The dataset contains $10^{4}$ images. We use this dataset solely for final evaluation of trained models.
\end{LaTeXdescription}
Illustrative examples of each dataset are shown in Figure~\ref{fig:datasetsamples}.

\subsection{BlendTorch Dataset}
\label{sec:btorchdataset}
The BlendTorch dataset is generated as follows. For each scene we randomly draw a fixed number of objects according to the current class probabilities which might change over the course of training. Each object is represented by its CAD model. For each generated object, there is a chance of producing an occluder. Occluding objects are based on super-shapes\footnote{\url{https://github.com/cheind/supershape}}, which can vary their shape significantly based on 12 parameters. These parameters are selected uniformly at random within a sensible range. Next, we randomize each objects position, rotation and procedural material. We ensure that initially each object hovers above a ground plane. Finally, we employ Blender's physics module to allow the objects to fall under the influence of gravity. Once the simulation has settled, we render $N$ images using a virtual camera positioned randomly on hemispheres with varying radius. Besides color images, we generate the following annotations: bounding boxes, class numbers and visibility scores. Figure~\ref{fig:datasetsamples:btorch} shows an example output. Each image and annotation pair is then published through BlendTorch data channels (refer to Figure~\ref{fig:arch} and Example~\ref{fig:minimalexample}).

\subsection{Training}
\label{sec:training}

Object detection is performed by CenterNet~\cite{zhou2020tracking, zhou2019objects} using a DLA-34~\cite{yu2018deep} feature extracting backbone pre-trained on ImageNet~\cite{deng2009imagenet}. The backbone features are inputs to multiple prediction heads. Each head consists of a convolution layer mapping to 256 feature channels, a rectified linear unit, and another convolution layer to have a task specific number of output channels. In particular, our network uses the following heads: \begin{inparaenum}[a)]
\item one center point heatmap per class of dimension $\scriptstyle 6 \times \lfloor H/4 \rfloor \times \lfloor W/4 \rfloor$, and 
\item a bounding box size regression head of dimension $\scriptstyle 2 \times \lfloor H/4 \rfloor \times \lfloor W/4 \rfloor.$
\end{inparaenum} 

During training, the total loss $L_{total}$ is computed as follows
\begin{equation}
    L_{total} = L_{hm} + 0.1 \cdot L_{wh}, \label{eq:loss}
\end{equation}
where $L_{hm}$ is measured as the focal loss of predicted and ground truth center point heatmaps, $L_{wh}$ is the L1 loss of regressed and true bounding box dimensions evaluated at ground truth center point locations.

We split the training dataset into training and validation data (\SI{90/10}{\percent}) and train for $2 \times 10^{5}$ steps using Adam~\cite{kingma2014adam} (lr=$1.25 \times 10^{-4}$, weight-decay=$10^{-3}$). We perform the following augmentations independent of the dataset: scale to \SI{512}{\px}, random rotation, random horizontal flip, and color jitter. Regularization and augmentation are applied to all training sets to avoid skewing of results due to different dataset sizes. 

Validation is performed every $1 \times 10^{5}$ steps. The best model is selected based on the lowest validation total loss, as defined in Equation~\ref{eq:loss}. As mentioned above, each training session is repeated 10 times, and the best model of each run is stored for evaluation.

\subsection{Prediction}

Objects and associated classes are determined by extracting the top-$K$ local peak values that exceed a given confidence score from all center heatmaps. Bounding box dimensions are determined from the respective channels at center point locations.

\subsection{Average Precision}
We assess the quality of each training dataset by computing the \acrfull{map}~\cite{everingham2010pascal} from model predictions based on the unseen BOP test dataset  (see Section~\ref{sec:datasets}) using the \acrfull{iou} evaluation metric. The additional training runs per model allow us to compute the \acrshort{map} with confidence. Unless otherwise stated, we report the \acrshort{map} averaged over runs, error bars indicate $\pm1$ standard deviation. We consider up to 25 model predictions that surpass a minimum confidence score of 0.1 during computation of the \acrshort{map}. 

\begin{figure}
    \centering
    \includegraphics[width=0.9\columnwidth]{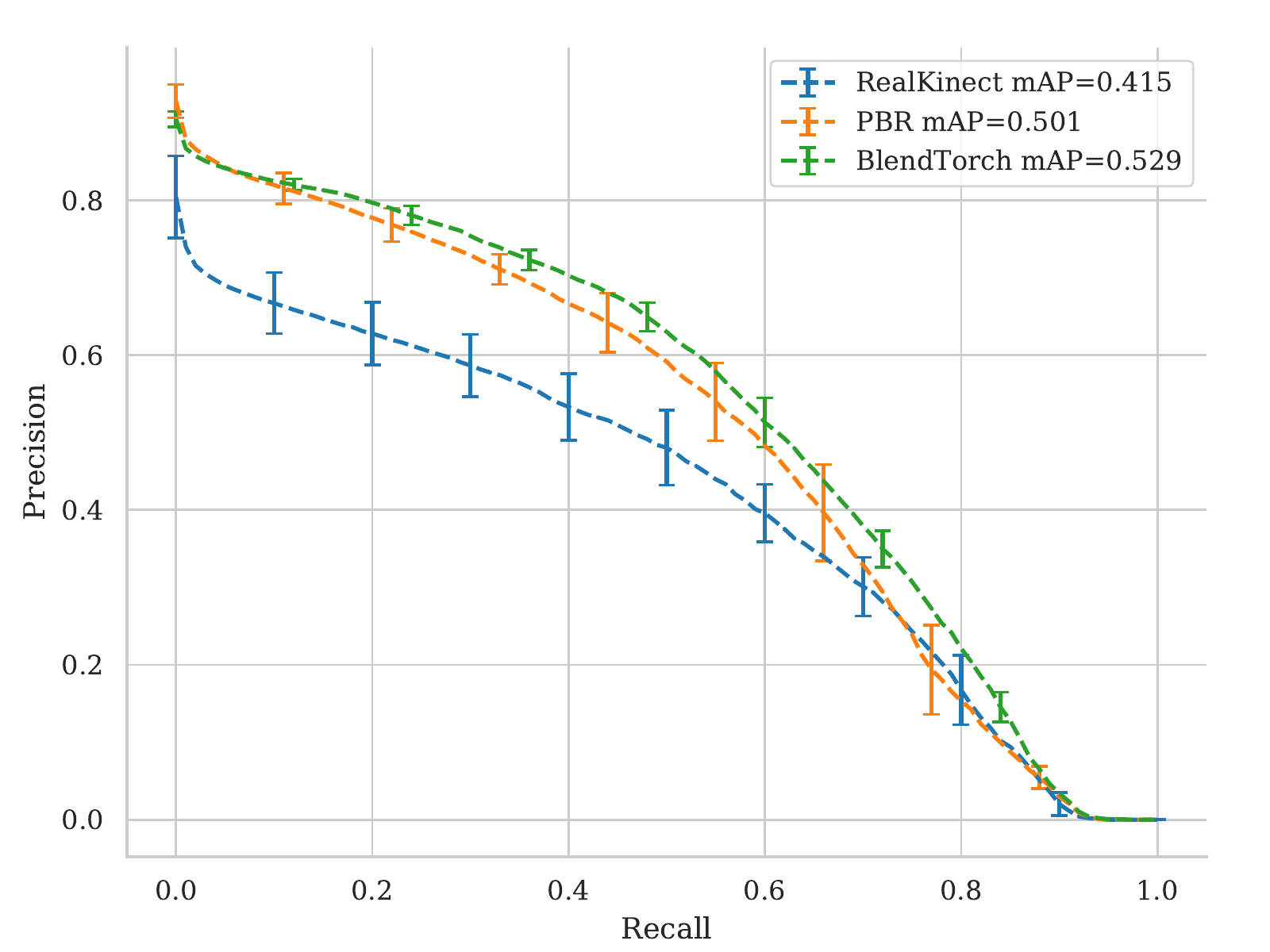}
    \caption{Precision-recall curves as a result of evaluating object detection performance on varying training datasets. Despite the non-realistic appearance of domain randomized scenes in BlendTorch, our approach outperforms photo-realistic datasets and real image dataset at all \acrshort{iou} thresholds. We repeat the process 10 times and indicate the variations as $\pm 1 \sigma$ error bars. Note that models trained with BlendTorch data have significantly less variability.}
    \label{fig:precpreccurves}
\end{figure}

Table~\ref{tab:apcompare} compares the average precision achieved for each training dataset over the course of 10 runs. BlendTorch outperforms photo-realistic and real image datasets. Although the BOP test dataset contains similar scenes to RealKinect, the characteristics of the cameras are different. Despite moderate model regularization, the network begins to overfit on the smaller RealKinect dataset, resulting in poor test performance. BlendTorch generated images only slightly exceed the data provided by the photo-realistic PBR data set. However, BlendTorch trained models exhibit only half the standard deviation of all other models, making their performance more predictable in real world applications. This feature is shown clearly in Figure~\ref{fig:precpreccurves} that compares the precision-recall behaviour of all three training datasets.

\begin{table*}[t]
\centering
\begin{tabular}{lcccclccccccc}
\hline
\addlinespace[0.5em]
           & \multicolumn{4}{c}{\textit{Overall AP}} &  & \multicolumn{6}{c}{\textit{Per-Class mAP}} \\ \cline{2-5} \cline{7-12} 
\addlinespace[0.5em]
\textit{Dataset}           & mAP & $\sigma_{\textrm{mAP}}$ & $\textrm{AP}_{50}$ & $\textrm{AP}_{75}$ &  & $\textrm{mAP}_{C0}$ & $\textrm{mAP}_{C1}$ & $\textrm{mAP}_{C2}$ & $\textrm{mAP}_{C3}$ & $\textrm{mAP}_{C4}$ & $\textrm{mAP}_{C5}$ \\ \hline
\addlinespace[0.5em]
RealKinect       & 41.1 & 3.4   &  71.7       &   42.9      &  & 29.9   & 52.5       &  37.0     & 47.3      &  28.4     & 53.7  \\
\addlinespace[0.5em]
PBR        & 50.1 & 2.9      &  70.8       &   58.8      &  & \bftab 40.3   & 49.0       &  \bftab 59.4     & 36.6      &  \bftab 59.2     & 56.0  \\
\addlinespace[0.5em]
BlendTorch (ours) & \bftab 52.3 & \bftab 1.2 & \bftab 73.5 &   \bftab 61.5      &  & 34.0   & \bftab 54.4       &  56.3     & \bftab 54.6      &  55.3     & \bftab 59.3  \\
\bottomrule
\end{tabular}%
\caption{Performance comparison. Average precision values over 10 training runs. Here $\textrm{mAP}$ refers to the mean average precision computed by integrating over all classes, \acrfull{iou} thresholds and all runs. $\sigma_{\textrm{mAP}}$ denotes standard deviation of $\textrm{mAP}$ measured over 10 runs. $\textrm{AP}_{50}$ and $\textrm{AP}_{75}$ represent average precision values for specific \acrshort{iou}-thresholds averaged over all runs and classes. Finally, $\textrm{mAP}_{Ci}$ are average precision values averaged for specific classes averaged over all runs and thresholds. Except for $\sigma_{\textrm{mAP}}$, higher values indicate better performance.}
\label{tab:apcompare}
\end{table*}

\subsection{Runtime}
We compare the time it takes to generate and receive synthetic images using BlendTorch. All experiments are performed on the same machine (Nvidia GeForce GTX 1080Ti) and software (Blender 2.90, PyTorch 1.60). Table~\ref{tab:runtime} shows the resulting runtimes image data synthesis. Compared to photo-realistic rendering, BlendTorch creates images at interactive frame rates, even for physics enabled scenes involving millions of vertices.
\begin{table}[h]
\centering
\begin{tabular}{@{}lccc@{}}
\toprule
                &           & \multicolumn{2}{c}{Scene {[}Hz{]}} \\ \cmidrule(l){3-4} 
Renderer        & Instances & Cube         & T-Less       \\ \midrule
BlendTorch (renderer Eevee) & 1         & 43.5        & 5.7       \\
                & 4         & 111.1        & 18.2       \\
Photo-realistic (PBR)          & 1         & 0.8        & 0.4       \\
                & 4         & 1.9        & 0.6 \\
\bottomrule
\end{tabular}
\caption{BlendTorch vs. photo-realistic rendering times. Timings are in frames per second. Higher values are better. All timings include the total time spent in rendering plus the time it takes to receive the data in training. We show timings for two different scenes, render engines and various numbers of parallel simulation instances. Cube represents a minimal scene having one object, whereas T-Less refers to a complex scene involving multiple objects, occluders and physics. Shared settings across all experiments: batch size (8), image size (\SI[product-units = single]{640 x 512 x 3}{\px}), and number of PyTorch workers (4).}
\label{tab:runtime}
\end{table}

\section{Conclusion}
We introduced BlendTorch, a real-time, adaptive \acrlong{dr} library for infinite artificial training data generation in industrial applications. Our object detection experiments show that---all other parameters being equal--- models trained with BlendTorch data outperform those models trained on photo-realistic or even real training datasets. Moreover, models trained with domain randomized data exhibit less performance variance over multiple runs. This makes \acrshort{dr} a useful technique to compensate for the lack of training data in industrial machine learning applications. For the future we plan to explore the possibilities of adjusting simulation parameters with respect to training progress and to apply BlendTorch in the medical field, which exhibits similar shortcomings of real training data.

\printbibliography

\end{document}